\documentclass[runningheads]{llncs}
\usepackage[T1]{fontenc}
\usepackage{graphicx}
\usepackage{hyperref}
\usepackage{color}

\usepackage[shortcuts,acronym]{glossaries}
\usepackage{booktabs}
\usepackage{subcaption}
\usepackage{amsmath}
\usepackage{amssymb}
\usepackage{algorithm}
\usepackage{algpseudocode}
\usepackage{numprint}
\usepackage{ifthen}
\usepackage{xifthen}
\usepackage{placeins}

\DeclareMathOperator*{\argmin}{arg\,min \;}
\DeclareMathOperator*{\argmax}{arg\,max \;}

\newcommand*{\name}[1]{\textsc{#1}}

\newcommand{\ie}{i.e.}
\newcommand{\eg}{e.g.}
\newrobustcmd{\textBF}[1]{\fontseries{b}\selectfont #1}
\newcommand{\ress}[2][]{%
    \ifthenelse{\isempty{#1}}
               {#2}
               {#2\(\,\pm\,\)#1}
} 
\newcommand{\bress}[2][]{%
    \ifthenelse{\isempty{#1}}
               {\textBF{#2}}
               {\textBF{#2}\(\,\pm\,\)\textBF{#1}}
}
\newcommand{\sress}[2][]{%
    \ifthenelse{\isempty{#1}}
               {\underline{#2}}
               {\underline{#2\(\,\pm\,\)#1}}
}
\newcommand{\pc}{\hphantom{0}}

\newacronym{DNN}{DNN}{deep neural network}
\newacronym{ARIMA}{ARIMA}{autoregressive integrated moving average}
\newacronym{AutoML}{AutoML}{automated machine learning}
\newacronym{AIC}{AIC}{Akaike information criterion}
\newacronym{BO}{BO}{Bayesian optimization}
\newacronym{CASH}{CASH}{combined algorithm selection and hyperparameter optimization}
\newacronym{DTW}{DTW}{dynamic time warping}
\newacronym{ES}{ES}{exponential smoothing}
\newacronym{GRU}{GRU}{gated recurrent units}
\newacronym{HPO}{HPO}{hyperparameter optimization}
\newacronym{KDE}{KDE}{kernel density estimation}
\newacronym{LSTM}{LSTM}{long-term short memory}
\newacronym{MASE}{MASE}{mean absolute scaled error}
\newacronym{ML}{ML}{machine learning}
\newacronym{RUL}{RUL}{remaining useful life}
\newacronym{RNN}{RNN}{recurrent neural network}
\newacronym{STL}{STL}{seasonal decomposition}
\newacronym{SVM}{SVM}{support vector machine}
\newacronym{SOTA}{SOTA}{state-of-the-art}

\begin{document}
\title{auto-sktime: Automated Time Series Forecasting}
%
%
\author{Marc-Andr\'e Z\"oller\inst{1,3}\orcidID{0000-0001-8705-9862} \and
Marius Lindauer\inst{2}\orcidID{0000-0002-9675-3175} \and
Marco F. Huber\inst{3,4}\orcidID{0000-0002-8250-2092}}
\authorrunning{M.-A. Z\"oller et al.}
%
\institute{USU GmbH, R\"uppurrer Str. 1, Karlsruhe, Germany
\email{mazoeller@gmail.com} \and
Institute of Artificial Intelligence, Leibniz University Hannover, Welfengarten 1, Hannover, Germany
\email{m.lindauer@ai.uni-hannover.de} \and
Institute of Industrial Manufacturing and Management IFF, University of Stuttgart \and
Center for Cyber Cognitive Intelligence CCI, Fraunhofer IPA, Nobelstr. 12, Stuttgart, Germany
\email{marco.huber@ieee.org}
}
\maketitle              
\begin{abstract}
In today's data-driven landscape, time series forecasting is pivotal in decision-making across various sectors. Yet, the proliferation of more diverse time series data, coupled with the expanding landscape of available forecasting methods, poses significant challenges for forecasters. To meet the growing demand for efficient forecasting, we introduce \name{auto-sktime}, a novel framework for automated time series forecasting. The proposed framework uses the power of \ac{AutoML} techniques to automate the creation of the entire forecasting pipeline. The framework employs \acl{BO} to automatically construct pipelines from statistical, \ac{ML} and \ac{DNN} models. Furthermore, we propose three essential improvements to adapt \ac{AutoML} to time series data. First, pipeline templates to account for the different supported forecasting models. Second, a novel warm-starting technique to start the optimization from prior optimization runs. Third, we adapt multi-fidelity optimizations to make them applicable to a search space containing statistical, \ac{ML} and \ac{DNN} models. Experimental results on 64 diverse real-world time series datasets demonstrate the effectiveness and efficiency of the framework, outperforming traditional methods while requiring minimal human involvement.

\end{abstract}

\section{Introduction}
\label{sec:introduction}
Time series forecasting is crucial for applications like economics, finance, and manufacturing. It involves analyzing historical data and using statistical, \acf{ML} or \acf{DNN} models to predict future trends. With accurate forecasts, businesses can efficiently allocate resources, plan production schedules, and manage inventories, reducing waste and costs. Moreover, time series forecasting can help manufacturers detecting potential issues before they become problematic. For example, predictive maintenance models can analyze sensor data to anticipate when a machine will need maintenance.

Due to the tedious and error-prone process of creating well-performing models \cite{Bischl23}, organizations struggle to create forecasting models. This issue is further amplified by a shortage of skilled data scientists \cite{Bauer2020}. This has led to an increasing demand for automation that can enable businesses to develop accurate forecasting models without requiring a high level of technical expertise. \Acf{AutoML} is an emerging field aiming to automate the process of creating, evaluating, and deploying \ac{ML} models. By leveraging heuristics and techniques from optimization theory, \ac{AutoML} tools search for the best models and hyperparameters to create accurate forecasts with minimal human interaction. Optimally, the \ac{AutoML} system provides a complete end-to-end automation covering steps like data cleaning, model selection, and hyperparameter tuning.

More formally, \ac{AutoML} aims at generating a pipeline, described by its hyperparameters \(\vec{\lambda}^\star\), minimizing
\begin{equation}
    \label{eq:automl}
    \vec{\lambda}^\star \in \argmin_{\vec{\lambda} \in \Lambda} \mathcal{L}(\mathcal{D}_{\mathrm{train}}, \mathcal{D}_{\mathrm{valid}}, \vec{\lambda})
\end{equation}
with \(\Lambda\) the configuration space containing all possible hyperparameter combinations, \(\mathcal{L}\) a validation loss function, and \(\mathcal{D}\) a dataset split into a train (\(\mathcal{D}_{\mathrm{train}}\)) and validation part (\(\mathcal{D}_{\mathrm{valid}}\)). In general, the features of Equation~\eqref{eq:automl} are not known as they depend on \(\mathcal{D}\), making the use of efficient solvers impossible.

While prior work on \ac{AutoML} mainly focused on classification and regression, \eg, \cite{Erickson2020,Feurer2015}, some work has been done to apply \ac{AutoML} techniques to time series forecasting, \eg, \cite{Deng2022,Shah2021,daSilva2022}. Yet, those methods mainly apply \ac{AutoML} procedures developed for tabular data to time series data. Consequently, the potential of \ac{AutoML} adapted to time series is currently not fully utilized. For example, multi-fidelity approximations, which are an integral part of \ac{AutoML} \cite{Hutter2018a}, usually use either a random subset of the training data or a number of fitting iterations as budget. Yet, both are not easily applicable to time series: random subsets distort the temporal relation of the data, and many forecasting models do not support iterative fitting. Therefore, we propose \name{auto-sktime}, an \ac{AutoML} framework for end-to-end time series forecasting. Our contributions can be summarized as follows:

\begin{enumerate}
    \item We propose \name{auto-sktime}, a framework combining statistical, \ac{ML}, and \ac{DNN} \ac{SOTA} techniques for time series forecasting, making it applicable to a wide range of time series. We use a \emph{templating} method to select appropriate pipelines given the input data. Each template encodes specific best practices to ensure an efficient yet flexible creation of pipelines.
    \item We propose a novel method for warm-starting the \ac{AutoML} optimization based on prior optimizations to increase the sampling efficiency of the optimization. While this method is specifically designed for time series, it can also be transferred to other \ac{AutoML} problems like classification.
    \item Current multi-fidelity optimization techniques are not applicable to common time series data. We propose a novel multi-fidelity budget enabling the benefits of multi-fidelity approximations for all kinds of time series data.
\end{enumerate}

The rest of this work is structured as follows: Section~\ref{sec:related_work} introduces related work in time series forecasting and \ac{AutoML}. The proposed \name{auto-sktime} framework is described in Section~\ref{sec:methods} and validated in Section~\ref{sec:experiments}. Finally, the results are discussed in Section~\ref{sec:conclusion}.

\section{Background and Related Work}
\label{sec:related_work}
First, we provide an introduction to existing forecasting models and techniques for automating time series forecasting. Finally, we also provide a short introduction to \acl{BO}.

\subsection{Models for Time Series Forecasting}
The first approaches for time series forecasting used statistical models like \ac{ARIMA}, \ac{ES}, and \ac{STL}. These models assume that future observations are related to past observations and can be predicted using the patterns observed in the data. The \ac{ARIMA} model \cite{Box2015}, for example, captures the autocorrelation in the data and uses it to predict future observations. \ac{ES} \cite{Hyndman2008b} assigns different weights to past observations, with more recent observations being given more weight. \ac{STL} models decompose the time series into seasonal, trend, and residual components \cite{Hyndman2021}. While statistical models offer interpretability and insights into the underlying dynamics of the data, they often struggle to handle complex relationships between variables \cite{Hyndman2021}. 

\ac{ML} models have gained significant attention in time series forecasting due to their ability to capture complex patterns and making fewer assumptions about the data. By modeling time series forecasts as a regression problem, supervised learning methods can be applied. Time series forecasting has seen extensive exploration of various \ac{ML} models, such as random forests \cite{Rady2021} or gradient boosting trees \cite{Januschowski2020}. \ac{ML} models offer flexibility, scalability, and the potential to handle diverse time series data with varying characteristics \cite{Masini2023}. However, achieving success with them often requires careful tuning of hyperparameters and access to large amounts of training data \cite{Januschowski2020}, which usually is not available if only a single time series is considered.

More recently, \acp{DNN} have emerged as powerful models for time series forecasting due to their ability to model temporal relations within the data. Specifically, \acp{RNN}, like \ac{LSTM} \cite{Hochreiter1997} and \ac{GRU} \cite{Cho2014a}, have shown remarkable performance in capturing long-term dependencies in sequential data \cite{Yamak2020}. Alternatively, transformers have been applied more recently to time series forecasting \cite{Zhou2021}. Additionally, feedforward networks can forecast time series by transforming the sequential input into a fixed-size feature representation \cite{Bontempi2013}. However, the successful application of neural networks for time series forecasting requires careful architecture design, appropriate activation functions, and regularization techniques. Ongoing research focuses on hybrid architectures, ensemble methods, and attention mechanisms to enhance the forecasting accuracy of \ac{DNN} models, \eg, \cite{Zhou2021}.

\subsection{Automated Time Series Forecasting}
Extensive studies have examined how to automate the creation and fine-tuning of statistical time series forecasting algorithms. For example, researchers have employed the Box-Jenkins methodology \cite{Box2015} to find optimized \ac{ARIMA} models while the state-space methodology \cite{Hyndman2008b} can be applied to \ac{ES} models. Yet, this prior work does not consider automating the selection of the underlying statistical model. The demand for automation in forecasting increases due to the various processes generating time series data and the ever-increasing number of models for time series forecasting (\ac{ML} and \ac{DNN} models).

While researchers have proposed various \ac{AutoML} approaches, the majority of these focus on standard learning tasks such as tabular or image classification, \eg, \cite{Erickson2020,Feurer2015}. These approaches typically treat the generation of \ac{ML} pipelines as a black-box optimization problem: Given a dataset \(\mathcal{D}\) and loss function \(\mathcal{L}\), \ac{AutoML} searches within a search space \(\Lambda\) to find a pipeline \(\vec{\lambda}^\star\) minimizing the validation loss. Researchers often address this optimization process using sample-efficient \ac{BO} \cite{Brochu2010}. Yet, time series forecasting, especially when forecasting single univariate time series, often has too little training data for complex regression models.

Multiple dedicated \ac{AutoML} frameworks for time series forecasting exist. \name{TSPO} \cite{Morten2020} transforms the forecasting problem into a tabular regression problem compatible with standard \ac{AutoML}. \name{Auto-PyTorch} \cite{Deng2022} constructs ensembles of \acp{DNN}. While it contains many standard features from \ac{AutoML}---like multi-fidelity approximations and ensemble learning---it confines itself to \acp{DNN}. \cite{daSilva2022} propose automatic forecasting of time series based on genetic optimization. By combining statistical and \ac{ML} models to a single search space, forecasting of time series panel data is possible. Similarly, \name{AutoTS} \cite{Catlin2020} also employs genetic optimization combined with creating an ensemble of evaluated candidates. \name{BOAT} \cite{Kurian2021} uses \ac{BO} to automatize the creation of statistical and \ac{ML} models. The optimization is warm-started using the typical \ac{AutoML} approach via meta-features of the given dataset. \name{HyperTS} \cite{Zheng2022} also uses genetic optimization of statistical and \ac{DNN} models combined with a greedy ensemble construction for univariate and multivariate time series. \name{AutoGluon-TS} \cite{Shchur2023} combines \acp{DNN}, \ac{ML}, and statistical models for forecasting. Finally, \name{AutoAI-TS} \cite{Shah2021} offers an end-to-end automation for forecasting of univariate and multivariate time series. However, it uses a fixed ensemble of diverse models and does not optimize an internal objective function as it is characteristic for \ac{AutoML}. Similarly, \name{PyAF} \cite{Carme2016} builds best-practice pipelines by combining different preprocessing steps with \ac{ML} regression models using their default hyperparameters.

While different approaches of \ac{AutoML} for forecasting exist, no framework combines all \ac{SOTA} performance improvements of \ac{AutoML} frameworks---namely ensembling, multi-fidelity approximations, and meta-learning. Frameworks that implement some of these improvements usually do not adapt them to the time series data format. With \name{auto-sktime}, we aim to overcome these limitations.

\subsection{Bayesian Optimization}
\acf{BO} \cite{Brochu2010} is often used to solve the \ac{AutoML} optimization problem described in Equation~\eqref{eq:automl}. It aims to find the global minimum of an unknown function \(f: \Lambda \rightarrow \mathbb{R}\) using an initial experiment design \(S_0 = \{(\vec{\lambda}_i, l_i) \}_{i = 1}^M\) followed by a sequential selection of new candidates \(\vec{\lambda}_{n+1}\) and a corresponding function evaluation \(l_{n+1} = f(\vec{\lambda}_{n+1})\) to generate \(S_{n+1} = S_n \cup \{(\vec{\lambda}_{n+1}, l_{n+1})\}\). After each new observation, a probabilistic surrogate model \(p(f, S_{1:n})\) of \(f\) is constructed. Prominent examples of \(p\) are Gaussian processes \cite{Rasmussen2006}, random forests \cite{Hutter2011}, or Bayesian neural networks \cite{Li2023}. \(p\) is used in combination with an acquisition function \(\alpha(\vec{\lambda}, S_{1:n})\) to select the most promising \(\vec{\lambda}_{n + 1}\) by trading off exploration and exploitation. By maximizing the acquisition function
\begin{equation}
    \label{eq:acquisiton_function}
    \vec{\lambda}_{n + 1}^\star \in \argmax_{\vec{\lambda} \in \Lambda} \alpha \left(\vec{\lambda}; S_{1:n}, p\right)
\end{equation}
in each iteration, a candidate \(\vec{\lambda}_{n + 1}^\star\) is selected for evaluation. Prominent examples of acquisition functions are knowledge gradients, entropy search, expected improvements, and upper confidence bound.

\section{\name{auto-sktime} Methodology}
\label{sec:methods}

\begin{figure*}[t]
    \centering
    \includegraphics[width=0.88\textwidth]{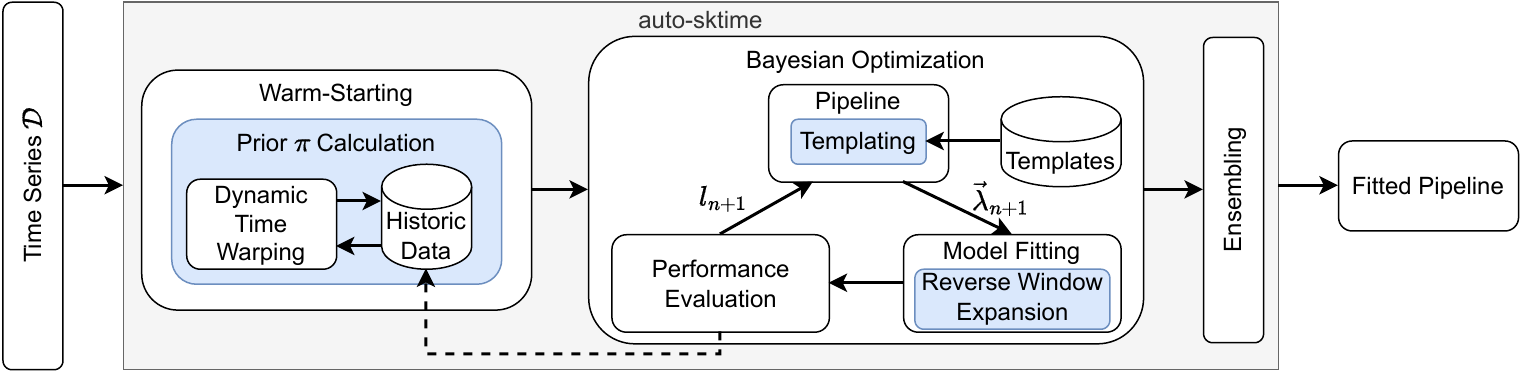}
    \caption{General architecture of \name{auto-sktime}. Highlighted in blue are our three proposed improvements for automatic time series forecasting.}
    \label{fig:architecture}
\end{figure*}

An overview of the proposed system architecture for \name{auto-sktime} is displayed in Figure~\ref{fig:architecture}. This architecture is similar to existing \ac{SOTA} \ac{AutoML} frameworks \cite{Erickson2020,Feurer2015}: Given the input data \(\mathcal{D}\), meta-learning is used to warm-start the \ac{BO}, more specifically \name{SMAC} \cite{Lindauer2022}. The optimization loop itself generates pipeline candidates \(\vec{\lambda}_{n + 1}\) from the set of predefined templates with corresponding hyperparameters. Candidates are fitted using multi-fidelity approximations to calculate their performance \(l_{n + 1}\). This loop is repeated until a predefined budget is exhausted. Finally, an ensemble is created using ensemble selection \cite{Caruana2004}. To adapt this methodology to time series forecasting, we need to introduce three new ideas, highlighted in blue in Figure~\ref{fig:architecture}, in more detail. Yet, first, we present a formal problem formulation for time series forecasting.

\subsection{Problem Formulation}
\label{sec:problem_formulation}
Let a dataset \(\mathcal{D} = \{D_i\}_{i = 1}^N\) contain \(N \in \mathbb{N}\) time series. Each time series \(D_i\) is defined by \(D_i = \{\vec{y}_{i, 1:T_i}, \vec{x}_{i, 1:T_i}^{(p)}, \vec{x}_{i, T_i+1:T_i+H}^{(f)}\}\) with \(T_i\) being the number of observations in \(D_i\) and \(H\) the forecasting horizon. Let \(\vec{y}_{i, 1:T_i}\), with \(\vec{y} \in \mathbb{R}^d\), denote the set of observed targets. In addition, let \(\vec{x}_{i, 1:T_i}^{(p)}\) and \(\vec{x}_{i, T_i+1:T_i+H}^{(f)}\), with \(\vec{x} \in \mathcal{X}^e\), denote the set of observed features and forecasted features, respectively. A forecasting model \(f_\mathcal{D}\) aims to predict future target values
\begin{equation}
\label{eq:forecasting}
    \vec{\hat{y}}_{i, T_i+1:T_i+H} = f_\mathcal{D} \Bigl( \vec{y}_{i, 1:T_i}, \vec{x}_{i, 1:T_i+H}; \vec{\lambda} \Bigl)
\end{equation}
with \(\vec{\lambda}\) being the hyperparameters describing the model. This definition covers univariate and multi-multivariate time series with or without exogenous data and even panel data containing multiple time series depending on the selected dimensions. The quality of the forecasts \(l \in \mathbb{R}\) is measured by the distance between the predicted and actual future targets according to a loss function \(\mathcal{L}\).

\subsection{Templates for Time Series}
\label{sec:templates}

Historically, three general approaches for time series forecasting have been proposed: statistical, \ac{ML}, and \ac{DNN} models. In general, it is not possible to simply focus on just one of these approaches as no single best approach exists \cite{Makridakis2020}, but the best approach depends on the actual dataset. For example, both \ac{ML} and \ac{DNN} methods are not applicable for short univariate time series, while statistical models are not able to generalize over panel data. In addition, as we aim to provide end-to-end automation for time series forecasting, just selecting the forecaster itself is not sufficient. Instead, multiple preprocessing steps for data cleaning and feature engineering are necessary. Yet, the actual pipeline steps depend on the selected forecasting method as different preprocessing is necessary for the three approaches. \ac{AutoML} tools often use a single fixed pipeline structure, \ie, \cite{Erickson2020,Feurer2015}, which is not able to capture the wide variety of potential input data, \ie, univariate/multivariate and endogenous/exogenous/panel data. Alternatively, \ac{AutoML} tools build arbitrary pipelines with genetic programming, \ie, \cite{DeSa2017}, which is not able to capture the implicit dependencies between the preprocessing steps and the used forecasting method.

We propose a templating approach to be used instead. Following the \acs{CASH} notation introduced by Thornton et al. \cite{Thornton2013} for a single pipeline template \(i\), the search space \(\Lambda_i\) aggregates a set of algorithms \(\mathcal{A}_i = \{A_i^1, \dots, A_i^n\}\) with according hyperparameters \(\Lambda_i^1, \dots, \Lambda_i^n\) into a single search space as \(\Lambda_i = \Lambda_i^1 \cup \dots \cup \Lambda_i^n \cup \{\lambda_{i, r}\}\), with \(\lambda_{i, r}\) being a hyperparameter selecting the algorithms in \(\mathcal{A}_i\). Instead of using a single pipeline template, we propose a joint search space \(\Lambda = \Lambda_1 \cup \dots \cup \Lambda_k \cup \{\lambda_r\}\), with \(\lambda_r\) selecting a pipeline, for \(k\) different templates.

\begin{figure*}[t]
    \centering
    \includegraphics[width=1.0\textwidth]{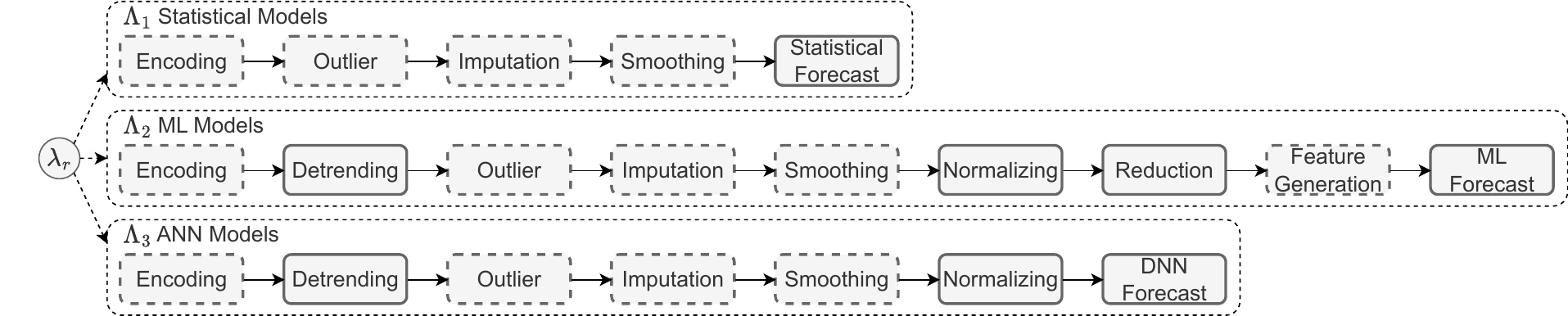}
    \caption{Overview of the available templates. Steps marked by a solid border are mandatory, steps marked by a dotted border are optional. For each step, multiple algorithms with corresponding hyperparameters are available.}
    \label{fig:templates}
\end{figure*}

Although our approach would allow for an arbitrary number of templates, we opted to implement three different pipeline templates, as there are three different classes of forecasting models with different requirements regarding data preprocessing. Each template is inspired by best-practice pipelines in the literature, \eg, \cite{Meisenbacher2022}. An overview of the templates is depicted in Figure~\ref{fig:templates}. While some steps are similar for all templates, \eg, encoding of categorical features, other steps only occur in a single template, like a reduction of time series data to tabular data for using \ac{ML} forecasters. These basic pipelines can be further expanded with more specialized steps (\eg, \cite{Candelieri2019}) in future work. No hard mapping of input data to a pipeline template exists. Meta-learning (see Section~\ref{sec:warm-starting}) can steer the optimization to a favorable combination of input data and template.


\subsection{Multi-Fidelity Approximation for Time Series}
\label{sec:multi-fidelity}

Creating a forecast using Equation~\eqref{eq:forecasting} can become expensive for sufficiently large input data \(\mathcal{D}\) as training a model with the according hyperparameters is necessary. Especially if many forecasts have to be created, as is the case with \ac{BO}, this procedure becomes quite expensive. Often, it is possible to define easier-to-evaluate proxies \(\tilde{f}_\mathcal{D}(\cdot, b)\) of \(f_\mathcal{D}(\cdot)\) that are parametrized by a budget \(b \in [b_\mathrm{min}, b_\mathrm{max}]\) with \(0 < b_\mathrm{min} < b_\mathrm{max} \leq 1\). These multi-fidelity approximations \(\tilde{f}_\mathcal{D}\) are used in \ac{AutoML} to speed up the evaluation of a single configuration \(\vec{\lambda}\) by discarding unpromising models early, \eg, \cite{Li2018}. The lowest budget \(b_\mathrm{min}\) is assigned to all initial models, and higher budgets are successively assigned to well-performing models until \(b_\mathrm{max}\) is reached and \(\tilde{f}(\cdot, b_\mathrm{max}) = f(\cdot)\).

To actually calculate \(\tilde{f}_\mathcal{D}\), a mapping of \(b\) to a cheaper function evaluation has to be defined. For tasks on tabular data often the number of training samples, \eg, \cite{Klein2016}, or number of iterations (\ie, epochs for neural networks or number of trees in random forests) are used, \eg, \cite{Bischl23}. For time series forecasting, these interpretations are not possible: many statistical models do not support iterative fitting. Selecting only some of the time series for training is only possible for panel data but not for univariate or multivariate forecasting. Similarly, reducing the length of a time series by selecting observations at random or selecting just every \(n\)-th sample is not always reasonable, as both methods could distort the seasonality of the data. Instead, we propose a reverse expanding window interpretation for multi-fidelity budgets.

\begin{figure}
    \centering
    \includegraphics[width=0.5\linewidth]{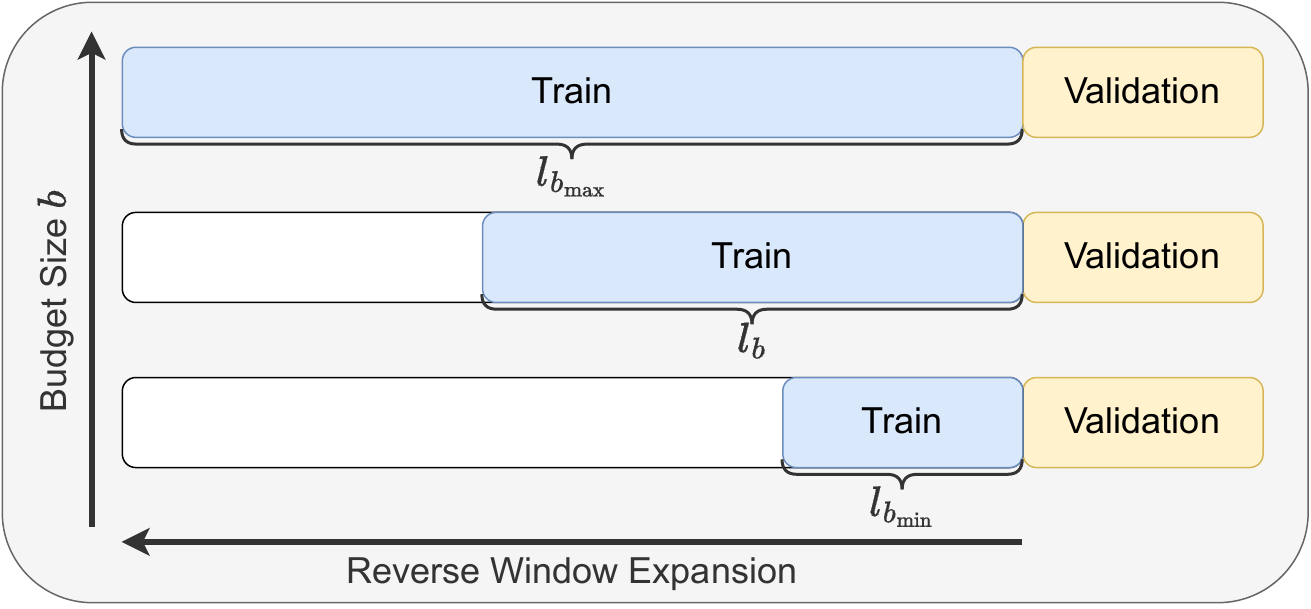}
    \caption{Interpretation of multi-fidelity budgets as reverse expanding windows.}
    \label{fig:multi-fidelity}
\end{figure}

The complete procedure of reverse expanding windows is visualized in Figure~\ref{fig:multi-fidelity}. Given the current budget \(b\), a desired window length \(l_b = \mathrm{min}(b \cdot T_i, l_\mathrm{min})\) is calculated. While multi-fidelity approximations can speed up the evaluation of forecasting models on long time series, the reduced computation time for short time series is only marginal. Therefore, they are deactivated if the time series is shorter than a user-defined constant \(l_\mathrm{min}\). Forecasts are created using only the latest \(l_b\) observations with \(\tilde{f}_\mathcal{D}\) being defined as
\begin{equation*}
    \vec{y}_{i, T_i+1:T_i+H} = \tilde{f}_\mathcal{D} \Bigl( \vec{y}_{i,T_i - l_b1:T_i}, \vec{x}_{i, T_i - l_b:T_i+H}; \vec{\lambda} \Bigl)~.
\end{equation*}
With increasing budgets, the window is expanding backward in time until, finally, the complete time series is included. The exact procedure to select budgets is handled using successive halving \cite{Li2018}.

\subsection{Warm-Starting the Optimization}
\label{sec:warm-starting}

In the context of \ac{AutoML}, warm-starting usually refers to the practice of using meta-learning to initialize the optimization process with prior knowledge about well-performing regions of the search space. This may lead to an accelerated optimization convergence and potentially better-performing models. Meta-learning, in the context of \ac{BO}, is commonly implemented by using specific configurations during initialization instead of a random initial design \(S_0\). Often, those configurations are a portfolio of well-performing configurations on a variety of datasets or similar datasets identified via meta-learning \cite{Vanschoren2019}.

Instead of using meta-learning only during the initialization and relying on the optimizer to derive well-performing regions during the optimization, Hvarfner et al. \cite{Hvarfner2022} proposed to include a user-defined prior \(\pi\) in the acquisition function, displayed in Equation~\eqref{eq:acquisiton_function}, to make prior information available
\begin{equation}
    \label{eq:decaying_prior_weighted_acq}
    \vec{\lambda}_n^\star \in \argmax_{\vec{\lambda} \in \Lambda} \alpha\left(\vec{\lambda}; S_{1:n}, p \right) \pi \left( \vec{\lambda} \right)^{\beta/n} 
\end{equation}
with \(\beta\) being a decay factor to fade out the prior in longer optimizations. Hvarfner et al. \cite{Hvarfner2022} were able to prove that the introduction of \(\pi ( \vec{\lambda} )^{\beta/n} \) does not negatively impact the worst-case convergence rate of \ac{BO} while showing improved performance empirically.

Equation~\eqref{eq:decaying_prior_weighted_acq} relies on user-provided priors \(\pi\) for each hyperparameter, which contradicts the end-to-end automation we pursue. Furthermore, providing reasonable priors, for instance, for all 98 hyperparameters included in \name{auto-sktime}, is virtually impossible for users. We propose to automatically extract priors from historic experiments to achieve a fully automated warm-starting instead of manually crafting priors. The proposed approach is outlined in Algorithm~\ref{alg:warm-starting}.

\begin{algorithm}[t]
\begin{algorithmic}[1]
    \Require Results of optimization runs on other datasets \(\mathcal{C}_\mathrm{meta}\), new time series \(\mathcal{D}_\mathrm{new}\)

    \For{\(\mathcal{D}_i \in \mathcal{C}_\mathrm{meta}\)} \label{alg:line:loop}
    \State    Calculate Distance \(d_i = d(\mathcal{D}_i, \mathcal{D}_\mathrm{new})\); Eq.~\eqref{eq:dist} \label{alg:line:d}
    \EndFor
    \State \(\mathcal{C}_\mathrm{sim}\) \(\leftarrow\) Select \(n_d\) closest datasets \label{alg:line:n_w}
    \State \(\Lambda' \subset \Lambda\) \(\leftarrow\) Concatenate all samples \(\{\vec{\lambda}\}\) from \(\mathcal{C}_\mathrm{sim}\) \label{alg:line:c}
    \State \((\Lambda', \vec{w})\) \(\leftarrow\) Weight \(\vec{\lambda} \in \Lambda'\) with \(d_i\) \label{alg:line:w}; see Eq. \eqref{eq:weight}
    \State \(\pi\) \(\leftarrow\) Fit \acs{KDE} using \(\{(\Lambda', \vec{w})\}\) \label{alg:line:kde}
\end{algorithmic}
\caption{Automatic prior calculation for warm-starting the forecasting.}
\label{alg:warm-starting}
\end{algorithm}

For each element of a set of historical time series, let a set of the \(n_c\) best historic hyperparameters be given as \(\mathcal{C}_\mathrm{meta} = \left\{ \left( \{\vec{\lambda}_j\}_{j = 1}^{n_c}, \mathcal{D}_i \right)\right\}_{i = 1}^k \). Furthermore, let a new dataset \(\mathcal{D}_{\mathrm{new}}\) be given that is going to be optimized. We aim to identify time series \(\mathcal{D}_i \in \mathcal{C}_{meta}\) with a low distance \(d: \mathcal{D} \times \mathcal{D} \rightarrow \mathbb{R}\) to \(\mathcal{D}_\mathrm{new}\)(Line~\ref{alg:line:loop}~and~\ref{alg:line:d}). Instead of using a distance based on hard-to-select meta-features, as it is usually done for tabular data, we propose to use native distance metrics for time series. In the context of this work, a distance
\begin{equation}
    d(\mathcal{E}, \mathcal{F}) = \dfrac{1}{M N} \sum_{i=1}^M \sum_{j=1}^N \mathrm{DTW}(E_{i, T_i-h:T_i}, F_{j, T_i-h:T_i} ) \label{eq:dist}
\end{equation}
based on \ac{DTW} \cite{Vintsyuk1968} is used with \(M\) and \(N\) being the number of time series in \(\mathcal{E}\) and \(\mathcal{F}\), respectively, and \(h\) a user-provided maximum sequence length. \ac{DTW} is used to compute the optimal alignment between two time series by finding the warping path that minimizes the cumulative distance between their data points. Similar distance measures would also be possible given that time series with different lengths are supported. To limit the introduced computational overhead for long time series, only the latest \(h\) observations of each time series are considered. Let \(\mathcal{C}_\mathrm{sim} = \{ ( \{\vec{\lambda}\}_i, \mathcal{D}_i ) \}_{i = 1}^{n_d} \subset \mathcal{C}_\mathrm{meta}\), with \(n_d\) being a user-provided hyperparameter, denote the set of time series most similar to \(\mathcal{D}_\mathrm{new}\) (Line~\ref{alg:line:n_w}) and \(\vec{d}_{n_d}\) the set of according distances. Next, all configurations selected in \(\mathcal{C}_\mathrm{sim}\) are combined \(\Lambda' = \bigcup_{\{\vec{\lambda}_j\}, \cdot \in \mathcal{C}_\mathrm{sim}} \{\vec{\lambda_j}\}\) (Line~\ref{alg:line:c}). Furthermore, the normalized distances
\begin{equation}
    \vec{w} = 1 - \dfrac{\vec{d}_{n_d} - \mathrm{min}(\vec{d}_{n_d})}{\mathrm{max}(\vec{d}_{n_d}) - \mathrm{min}(\vec{d}_{n_d})} \label{eq:weight}
\end{equation}
are used to weight \(\Lambda'\) to give higher importance to configurations tested on data very similar to \(\mathcal{D}_\mathrm{new}\). Finally, based on \(\Lambda'\) and \(\vec{w}\) we construct the prior \(\pi\) using \ac{KDE} \cite{Chen2017} (Line~\ref{alg:line:kde}), modeling promising areas of the configuration space. As \cite{Pushak2022} were able to show that there are only a few interactions between the different hyperparameters in \(\vec{\Lambda}\), we also model their priors independently using univariate \acp{KDE}. The performance of configurations in \(\Lambda'\) is not further considered, but only the distribution of \(\Lambda'\) is relevant.


\section{Experiments}
\label{sec:experiments}

To show the viability of \name{auto-sktime}, we perform a thorough evaluation on 64 real-world datasets. \name{auto-sktime} is compared with four \ac{SOTA} forecasting frameworks, namely \name{PMDArima} \cite{Smith2017}, \name{ETS} \cite{Hyndman2008b}, \name{DeepAR} \cite{Salinas2020}, and \name{TFT} \cite{Lim2019}, as a baseline. Additionally, we also test the aforementioned frameworks already employing \ac{AutoML} techniques which have published source code, namely, \name{Auto-PyTorch} (APT-TS), \name{AutoTS}, \name{HyperTS}, \name{PyAF}, and \name{AutoGluon-TS}. In contrast to benchmarks used in the related work, we consider a mix of univariate, multivariate, and panel data. Furthermore, we focus on rather short time series as they are more prevalent in real-world applications \cite{Vogelsgesang2018}. Each benchmark dataset comes with a predefined forecasting horizon that is used as a holdout test set. In most cases, this forecasting horizon corresponds to one natural period, \eg, 24 observations for hourly data. The performance of each framework is measured using \ac{MASE}. All evaluations are repeated five times with different initial seeds to account for non-determinism. Each evaluation was limited to a computational budget of \(5\) minutes. Evaluations still running after an additional grace period of \(60\) seconds were pruned with the worst result produced by any of the competitors on the same dataset with an additional small failure penalty. While these optimization durations are rather short for typical \ac{AutoML} applications, preliminary results showed fitting time series forecasting models on the rather short datasets selected for the benchmark often required only a few seconds, and the used \ac{DNN} models converged in over 80\% of all evaluations before hitting the time limit. The priors for warm-starting are generated using a leave-one-out procedure by only considering the remaining 63 datasets for each evaluated dataset to prevent leaking knowledge. All frameworks are used with their default parameters to emphasize the idea of end-to-end optimization. The source code, scripts for the experiments, and detailed results are available on \name{Github} to ensure the reproducibility of all presented results.\footnote{
    See \url{https://github.com/Ennosigaeon/auto-sktime}.
}

\subsection{Datasets}


We conducted experiments using 64 real-world datasets sourced from different domains---including, for example, social media, finance, and commerce---encompassing both univariate, multivariate, and panel time series that are publicly available \cite{Carme2023,Maciag2021,Mulla2018}. Due to the lack of common time series forecasting benchmark datasets, we reused datasets used in related work. None of the selected datasets were used during the development of \name{auto-sktime}. The number of data samples in the univariate time series ranges from 144 to \numprint{145366}, while the multivariate time series, with up to \numprint{111} dimensions, range from \numprint{48} to \numprint{7588} samples. Panel datasets contain 32 to \numprint{1428} time series with 20 to 942 samples. The list of used datasets, with results, is available in the supplementary material.

\subsection{Experiment Results}
\label{sec:experiment_results}

Table~\ref{tab:results} shows the final test performance of all evaluations. For each framework, the mean \ac{MASE} score, ranking, and fitting time in seconds with the according standard deviations are given. Bold face represents the best mean value. Results not significantly worse, according to a \(t\)-test with \(\alpha = .05\) and Bonferroni correction, are underlined. \name{auto-sktime} significantly outperforms all other frameworks both regarding to the absolute performance (\ac{MASE}) and rank. Surprisingly, \name{ETS} and \name{pmdarima} reached a better average \ac{MASE} than some of the \ac{AutoML} tools. This can be explained as the \ac{AutoML} tools produced bad results on a minor subset of the dataset while \name{ETS} and \name{pmdarima} significantly outperform them on very few data sets (see Table~1 in the online appendix). When considering the median performance, the \ac{AutoML} tools outperform the baseline methods. This is also reflected in the average ranking in Table~\ref{tab:results}. We consider this a limitation of our small benchmark rather than a general issue of \ac{AutoML} tools. \name{TFT} and \name{DeepAR} perform badly due to failures on many time series. If a prediction was created, they often have a competitive performance.

\begin{table}[t]
    \centering
    \captionof{table}{Performance overview on all time series while enforcing timeouts.}
    \label{tab:results}
    
    \setlength{\tabcolsep}{3.5pt}
    \small
    
    \begin{tabular}{@{} l c c c @{}}
        \toprule
        Framework           & \ac{MASE}              & Ranking            & Time  \\
        \midrule
        \name{APT-TS}       & \ress[5.41]{3.12}      & \ress[2.49]{4.51}  & \ress[134.9]{242.2}  \\
        \name{auto-sktime}  & \bress[1.61]{1.59}     & \bress[1.94]{3.22} & \ress[43.4]{326.1}\pc  \\
        \name{AutoGluon}    & \pc\ress[18.17]{6.00}  & \ress[2.34]{6.74}  & \ress[160.1]{144.9}  \\
        \name{AutoTS}       & \pc\sress[10.94]{4.26} & \ress[2.35]{5.38}  & \ress[52.7]{321.1}\pc  \\
        \name{DeepAR}       & \ress[51.51]{11.54}    & \ress[1.99]{7.23}  & \pc\ress[106.0]{61.3}  \\
        \name{ETS}          & \ress[2.78]{2.57}      & \ress[2.34]{5.82}  & \pc\bress[7.6]{2.0}\pc  \\
        \name{HyperTS}      & \ress[5.03]{2.88}      & \ress[2.60]{4.57}  & \ress[89.4]{300.0}\pc  \\
        \name{pmdarima}     & \ress[3.21]{2.63}      & \ress[2.62]{5.42}  & \pc\ress[98.4]{47.1}\pc  \\
        \name{PyAF}         & \ress[5.33]{3.23}      & \ress[2.56]{5.38}  & \pc\ress[56.5]{22.6}\pc  \\
        \name{TFT}          & \pc\ress[50.62]{9.80}  & \ress[1.95]{6.73}  & \pc\ress[118.3]{76.9} \\
        \bottomrule
    \end{tabular}
\end{table}

A major issue for many \ac{AutoML} tools is the missing support for end-to-end automation of time series forecasting. Multiple frameworks are missing methods for data cleaning, \eg, imputation of missing values. As \(27\) time series contained missing values, the average performance of those frameworks appears to be significantly worse. Table~\ref{tab:results_without_missing_values}
contains the results of all tested frameworks on only time series not necessarily requiring preprocessing and data cleaning. Most methods achieve a better performance. As a consequence, \name{auto-sktime} does not significantly outperform many of the baseline methods anymore, even though it still has the best average performance. Interestingly, methods like \name{DeepAR} and \name{TFT} even profited from failing runs, as the worst performance of the competitors, on average, is still better than their own predictions.

\begin{table}[t]
    \centering
    \captionof{table}{Performance overview for two different scenarios.}
    \label{tab:results_without_missing_values}

    \setlength{\tabcolsep}{0.4pt}
    \small
    
    \begin{tabular}{@{} l c c c | c c c @{}}
        \toprule
                            & \multicolumn{3}{c|}{Without Missing Values}                        &  \multicolumn{3}{c}{Without Enforcing Timeouts} \\
        Framework           & \ac{MASE}              & Ranking            & Time                 & \ac{MASE}             & Ranking          & Time\\
        \midrule
        \name{APT-TS}       & \ress[6.1]{2.4}      & \ress[2.8]{4.7}  & \ress[109.2]{276.5}     & \ress[6.3]{3.4}      & \ress[2.5]{5.1}  & \ress[141.6]{246.3}  \\
        \name{auto-sktime}  & \bress[0.9]{1.0}     & \bress[1.8]{3.5} & \ress[12.4]{327.7}\pc   & \bress[1.6]{1.4}     & \bress[1.8]{2.9} & \ress[54.3]{334.6}\pc \\
        \name{AutoGluon}    & \pc\ress[23.4]{6.9}  & \sress[3.0]{6.7} & \ress[144.9]{224.9}     & \ress[4.7]{2.8}      & \ress[2.3]{5.2}  & \pc\ress[1480.3]{585.3}  \\
        \name{AutoTS}       & \pc\sress[14.0]{4.6} & \ress[2.0]{6.3}  & \ress[29.4]{318.7}\pc   & \pc\sress[10.5]{3.3} & \ress[2.5]{4.8}  & \ress[421.4]{512.5}  \\
        \name{DeepAR}       & \ress[66.7]{16.3}    & \ress[2.5]{7.5}  & \ress[121.5]{103.0}     & \ress[51.5]{11.9}    & \ress[1.8]{7.7}  & \pc\ress[106.0]{61.3} \\
        \name{ETS}          & \ress[1.1]{1.4}      & \ress[2.8]{5.3}  & \pc\bress[0.7]{0.3}\pc  & \ress[2.9]{2.7}      & \ress[2.2]{6.3}  & \bress[7.6]{2.0} \\
        \name{HyperTS}      & \ress[5.9]{2.4}      & \ress[2.5]{5.3}  & \ress[54.3]{311.0}\pc   & \ress[5.8]{2.5}      & \ress[2.8]{4.0} & \ress[155.5]{342.0}  \\
        \name{pmdarima}     & \sress[1.1]{1.2}     & \ress[3.1]{4.5}  & \ress[86.0]{51.3}       & \ress[4.6]{2.9}      & \ress[2.6]{5.8}  & \pc\ress[291.3]{88.6}  \\
        \name{PyAF}         & \ress[5.9]{2.3}      & \sress[2.9]{4.4} & \ress[70.1]{36.4}       & \ress[6.3]{3.6}      & \ress[2.4]{5.9}  & \ress[61.6]{23.4} \\
        \name{TFT}          & \sress[65.7]{13.3}   & \ress[2.5]{6.7}  & \ress[130.1]{129.1}     & \ress[50.7]{10.2}    & \ress[1.8]{7.2}  & \pc\ress[118.2]{76.9}   \\
        \bottomrule
    \end{tabular}
\end{table}

\begin{table}[t]
    \centering
    \captionof{table}{Performance of \name{auto-sktime} variants.}
    \label{tab:results_ablation_study}

    \setlength{\tabcolsep}{1.3pt}
    \small
    
    \begin{tabular}{@{} l c c c @{}}
        \toprule
        Framework               & \ac{MASE}           & Ranking            & Time  \\
        \midrule
        \name{APT-TS}           & \ress[3.45]{2.67}   & \ress[1.61]{3.44}  & \bress[134.9]{242.2}  \\
        \midrule
        \name{auto-sktime}      & \bress[1.60]{1.54}  & \bress[1.20]{2.23} & \ress[43.4]{326.1}\pc  \\
        \name{Templates}        & \ress[9.02]{3.21}  & \ress[1.15]{3.41}  & \ress[43.4]{324.9}\pc  \\
        \name{Templates \& Multi-Fidelity}   & \sress[2.22]{1.99} & \ress[1.26]{3.19}  & \ress[44.3]{326.7}\pc  \\
        \name{Templates \& Warm Starting}    & \sress[3.12]{2.16} & \ress[1.24]{2.73}  & \ress[44.0]{327.5}\pc  \\
        \bottomrule
    \end{tabular}
\end{table}

Another major issue was the imposed time limit of \(300\) seconds. Even though a grace period of additional \(60\) seconds was introduced, many tools did not adhere to the maximum budget with single optimization runs requiring over \numprint{5000} seconds. In total, \(10.36\)\% of all evaluations violated the timeout with \name{AutoTS} and \name{AutoGluon} exceeding it on nearly \(33.3\)\% of all runs worsening the reported results significantly. Table~\ref{tab:results_without_missing_values} summarizes the average performance of all tested frameworks when the timeout is not enforced. \name{AutoTS} and \name{AutoGluon} gained a significant performance boost as they regularly exceeded the configured timeout on larger time series. Similarly, \name{HyperTS}'s and \name{auto-sktime}'s performance also slightly improved. The remaining frameworks basically never utilized the configured optimization budget fully and consequently also did not improve their performance. Even so \name{ETS} only achieves a mediocre ranking, it is still able to beat some of the contestants in a fraction of the computational budget, proving the value of considering statistical forecasting methods.

In both examined alternative scenarios, \name{auto-sktime} still obtains the best average performance even though differences are often not significant anymore. Yet, in all scenarios, \name{auto-sktime} still achieves the overall best ranking.

\subsection{Ablation Study}
In Section~\ref{sec:templates}--\ref{sec:warm-starting}, we proposed three potential \ac{AutoML} improvements for time series forecasting. To study their impact, we perform an ablation study by introducing three variants of \name{auto-sktime}: \name{Templates} using only the templating approach, \name{Templates \& Multi-Fidelity} using templating and multi-fidelity approximations but not warm-starting, and \name{Templates \& Warm-Starting} using templating and warm-starting but not multi-fidelity. These three variants are evaluated on the same datasets as the other forecasting frameworks and compared with the original \name{auto-sktime} version, including all proposed improvements. We also provide the performance of \name{Auto-PyTorch}, the second-best method, as a reference for assessing the impact of the different versions.

As shown in Table~\ref{tab:results_ablation_study}, the templating base version performs worse than \name{Auto-PyTorch}. Adding either multi-fidelity approximations or warm-starting improves the performance significantly, with both versions outperforming \name{Auto-PyTorch}. By combining all three improvements, the performance of \name{auto-sktime} is again slightly improved, making all three proposed improvements useful for time series forecasting.

\section{Conclusion and Limitation}
\label{sec:conclusion}
By adapting existing \ac{AutoML} techniques to the domain of time series forecasting, instead of just applying techniques developed for tabular data, we were able to outperform existing \ac{AutoML} frameworks for time series forecasting. \name{auto-sktime} proved to be viable for forecasting on a wide variety of time series data, \ie, univariate, multivariate, and panel data with or without exogenous data. The ablation study showed that a combination of at least two proposed adaptations of \ac{AutoML} techniques for time series forecasting are necessary to outperform the current \ac{SOTA} while using templating, multi-fidelity approximations and warm-starting together yields an even better performance.

During the experiments, we were able to show that a robust handling of common data errors is important for good performance. Yet, while we strove for end-to-end automation of forecasting, \name{auto-sktime} is still not applicable to time series with arbitrary data defects. For example, \name{auto-sktime} is currently not able to handle measurements collected with an irregular frequency. We plan to overcome this and similar common data defects in the near future.

Besides forecasting, time series are also relevant for other learning tasks like, for example, time series classification or regression. While the presented framework was already used successfully for time series regression in the context of remaining useful life predictions of mechanical systems \cite{Zoller2023}, further work to validate the usefulness of \name{auto-sktime} on a variety of domains is necessary.

\subsubsection*{Acknowledgement}
Marc Z\"oller was funded by the Federal Ministry for Economic Affairs and Climate Action in the project AutoQML, Marius Lindauer by the Federal Ministry for Economic Affairs and Energy of Germany in the project CoyPu (Grant 01MK21007L), and Marco Huber by the Baden-Wuerttemberg Ministry for Economic Affairs, Labour and Tourism in the project KI-Fortschritts\-zentrum ``Lernende Systeme und Kognitive Robotik'' (Grant 036-140100).

\bibliographystyle{splncs04}
\bibliography{library}

\end{document}


\onecolumn
\aistatstitle{auto-sktime: Automated Time Series Forecasting\\Supplementary Materials}

\section{Raw Experiment Results}

\begin{table}[ht]
    \centering

    \tiny
    \setlength\tabcolsep{3.0pt}

    \caption{Raw experiment results for all evaluated frameworks and used datasets using MASE.}
    \label{table:raw-results}
    
\begin{tabular}{@{} l  c c  c c  c c  c c  c c  c c  c c  c c  c c  c c @{}}
    \toprule
Dataset & \multicolumn{2}{c}{Auto-PyTorch} & \multicolumn{2}{c}{auto-sktime} & \multicolumn{2}{c}{AutoGluon} & \multicolumn{2}{c}{AutoTS} & \multicolumn{2}{c}{DeepAR} & \multicolumn{2}{c}{ETS} & \multicolumn{2}{c}{HyperTS} & \multicolumn{2}{c}{pmdarima} & \multicolumn{2}{c}{PyAF} & \multicolumn{2}{c}{TFT} \\
        & \rotatebox[origin=c]{90}{MASE}&\rotatebox[origin=c]{90}{Duration} & \rotatebox[origin=c]{90}{MASE}&\rotatebox[origin=c]{90}{Duration} & \rotatebox[origin=c]{90}{MASE}&\rotatebox[origin=c]{90}{Duration} & \rotatebox[origin=c]{90}{MASE}&\rotatebox[origin=c]{90}{Duration} & \rotatebox[origin=c]{90}{MASE}&\rotatebox[origin=c]{90}{Duration} & \rotatebox[origin=c]{90}{MASE}&\rotatebox[origin=c]{90}{Duration} & \rotatebox[origin=c]{90}{MASE}&\rotatebox[origin=c]{90}{Duration} & 
        \rotatebox[origin=c]{90}{MASE}&\rotatebox[origin=c]{90}{Duration} & 
        \rotatebox[origin=c]{90}{MASE}&\rotatebox[origin=c]{90}{Duration} & \rotatebox[origin=c]{90}{MASE}&\rotatebox[origin=c]{90}{Duration} \\
\midrule

AEDemand                       &  --    &  310  &  0.18  &  320  &  1.09  &  100  &  1.11  &  308  &  1.04  &  6  &  0.94  &  0  &  1.00  &  305  &  0.00  &  1  &  1.15  &  21  &  1.07  &  197 \\
AEP\_hourly                    &  5.59  &  317  &  3.24  &  341  &  --    &  0  &  2.97  &  1256  &  --    &  0  &  --    &  0  &  3.05  &  563  &  --    &  0  &  --    &  0  &  --    &  0 \\
AirPassengers                  &  0.78  &  321  &  0.93  &  323  &  0.59  &  99  &  1.54  &  310  &  3.68  &  3  &  3.16  &  0  &  0.82  &  312  &  1.86  &  1  &  3.00  &  4  &  2.79  &  4 \\
DAYTON\_hourly                 &  5.81  &  370  &  3.91  &  327  &  --    &  0  &  3.20  &  1198  &  --    &  0  &  --    &  0  &  3.13  &  571  &  --    &  0  &  --    &  0  &  --    &  0 \\
\bottomrule
    \end{tabular}
\end{table}

\begin{table}[ht]
    \centering

    \tiny
    \setlength\tabcolsep{1.3pt}

    \caption{Raw experiment results for all evaluated frameworks and used datasets using RMSE. Predictions with consistent failures are marked with --.}
    
    \begin{tabular}{@{} l  c c  c c  c c  c c  c c  c c  c c  c c  c c  c c @{}}
    \toprule
Dataset & \multicolumn{2}{c}{Auto-PyTorch} & \multicolumn{2}{c}{auto-sktime} & \multicolumn{2}{c}{AutoGluon} & \multicolumn{2}{c}{AutoTS} & \multicolumn{2}{c}{DeepAR} & \multicolumn{2}{c}{ETS} & \multicolumn{2}{c}{HyperTS} & \multicolumn{2}{c}{pmdarima} & \multicolumn{2}{c}{PyAF} & \multicolumn{2}{c}{TFT} \\
        & \rotatebox[origin=c]{90}{RMSE}&\rotatebox[origin=c]{90}{Duration} & \rotatebox[origin=c]{90}{RMSE}&\rotatebox[origin=c]{90}{Duration} & \rotatebox[origin=c]{90}{RMSE}&\rotatebox[origin=c]{90}{Duration} & \rotatebox[origin=c]{90}{RMSE}&\rotatebox[origin=c]{90}{Duration} & \rotatebox[origin=c]{90}{RMSE}&\rotatebox[origin=c]{90}{Duration} & \rotatebox[origin=c]{90}{RMSE}&\rotatebox[origin=c]{90}{Duration} & \rotatebox[origin=c]{90}{RMSE}&\rotatebox[origin=c]{90}{Duration} & 
        \rotatebox[origin=c]{90}{RMSE}&\rotatebox[origin=c]{90}{Duration} & 
        \rotatebox[origin=c]{90}{RMSE}&\rotatebox[origin=c]{90}{Duration} & \rotatebox[origin=c]{90}{RMSE}&\rotatebox[origin=c]{90}{Duration} \\
\midrule

AEDemand                       &  --    &  310  &  0.48  &  320  &  2.51  &  100  &  2.45  &  308  &  2.22  &  6  &  2.26  &  0  &  2.41  &  305  &  0.00  &  1  &  2.28  &  21  &  2.50  &  197 \\
AEP\_hourly                    &  2807.39  &  317  &  1758.63  &  341  &  --    &  0  &  1526.48  &  1256  &  --    &  0  &  --    &  0  &  1517.44  &  563  &  --    &  0  &  --    &  0  &  --    &  0 \\
AirPassengers                  &  24.15  &  321  &  26.08  &  323  &  18.05  &  99  &  44.83  &  310  &  111.65  &  3  &  102.98  &  0  &  22.84  &  312  &  53.29  &  1  &  79.36  &  4  &  91.43  &  4 \\
DAYTON\_hourly                 &  442.30  &  370  &  317.72  &  327  &  --    &  0  &  242.41  &  1198  &  --    &  0  &  --    &  0  &  247.90  &  571  &  --    &  0  &  --    &  0  &  --    &  0 \\
\bottomrule
    \end{tabular}
\end{table}

\begin{table}[ht]
    \centering

    \tiny
    \setlength\tabcolsep{3.2pt}

    \caption{Raw experiment results for all evaluated frameworks and used datasets using sMAPE. Predictions with consistent failures are marked with --.}
    
    \begin{tabular}{@{} l  c c  c c  c c  c c  c c  c c  c c  c c  c c  c c @{}}
    \toprule
Dataset & \multicolumn{2}{c}{Auto-PyTorch} & \multicolumn{2}{c}{auto-sktime} & \multicolumn{2}{c}{AutoGluon} & \multicolumn{2}{c}{AutoTS} & \multicolumn{2}{c}{DeepAR} & \multicolumn{2}{c}{ETS} & \multicolumn{2}{c}{HyperTS} & \multicolumn{2}{c}{pmdarima} & \multicolumn{2}{c}{PyAF} & \multicolumn{2}{c}{TFT} \\
        & \rotatebox[origin=c]{90}{sMAPE}&\rotatebox[origin=c]{90}{Duration} & \rotatebox[origin=c]{90}{sMAPE}&\rotatebox[origin=c]{90}{Duration} & \rotatebox[origin=c]{90}{sMAPE}&\rotatebox[origin=c]{90}{Duration} & \rotatebox[origin=c]{90}{sMAPE}&\rotatebox[origin=c]{90}{Duration} & \rotatebox[origin=c]{90}{sMAPE}&\rotatebox[origin=c]{90}{Duration} & \rotatebox[origin=c]{90}{sMAPE}&\rotatebox[origin=c]{90}{Duration} & \rotatebox[origin=c]{90}{sMAPE}&\rotatebox[origin=c]{90}{Duration} & 
        \rotatebox[origin=c]{90}{sMAPE}&\rotatebox[origin=c]{90}{Duration} & 
        \rotatebox[origin=c]{90}{sMAPE}&\rotatebox[origin=c]{90}{Duration} & \rotatebox[origin=c]{90}{sMAPE}&\rotatebox[origin=c]{90}{Duration} \\
\midrule

AEDemand                       &  --    &  310  &  0.00  &  320  &  0.02  &  100  &  0.02  &  308  &  0.02  &  6  &  0.02  &  0  &  0.02  &  305  &  0.00  &  1  &  0.02  &  21  &  0.02  &  197 \\
AEP\_hourly                    &  0.16  &  317  &  0.09  &  341  &  --    &  0  &  0.08  &  1256  &  --    &  0  &  --    &  0  &  0.09  &  563  &  --    &  0  &  --    &  0  &  --    &  0 \\
AirPassengers                  &  0.04  &  321  &  0.05  &  323  &  0.03  &  99  &  0.08  &  310  &  0.19  &  3  &  0.16  &  0  &  0.04  &  312  &  0.09  &  1  &  0.16  &  4  &  0.14  &  4 \\
DAYTON\_hourly                 &  0.18  &  370  &  0.12  &  327  &  --    &  0  &  0.10  &  1198  &  --    &  0  &  --    &  0  &  0.10  &  571  &  --    &  0  &  --    &  0  &  --    &  0 \\
\bottomrule
    \end{tabular}
\end{table}

\begin{table}[ht]
    \centering

    \scriptsize
    \setlength\tabcolsep{3.5pt}

    \caption{Raw experiment results for all evaluated frameworks in the ablation study and used datasets using MASE. Predictions with consistent failures are marked with --.}
    
    \begin{tabular}{@{} l  c c  c c  c c  c c  c c @{}}
    \toprule
Dataset & \multicolumn{2}{c}{Auto-PyTorch} & \multicolumn{2}{c}{Multi-Fidelity} & \multicolumn{2}{c}{Templates} & \multicolumn{2}{c}{Warm-Starting} & \multicolumn{2}{c}{auto-sktime} \\
        & \rotatebox[origin=c]{90}{MASE}&\rotatebox[origin=c]{90}{Duration} & \rotatebox[origin=c]{90}{MASE}&\rotatebox[origin=c]{90}{Duration} & \rotatebox[origin=c]{90}{MASE}&\rotatebox[origin=c]{90}{Duration} & \rotatebox[origin=c]{90}{MASE}&\rotatebox[origin=c]{90}{Duration} & \rotatebox[origin=c]{90}{MASE}&\rotatebox[origin=c]{90}{Duration} \\
\midrule

AEDemand                       &  --    &  310  &  0.18  &  320  &  0.18  &  311  &  0.20  &  309  &  0.02  &  307 \\
AEP\_hourly                    &  5.59  &  317  &  3.24  &  341  &  3.24  &  416  &  2.31  &  320  &  2.31  &  323 \\
AirPassengers                  &  0.78  &  321  &  0.93  &  323  &  0.94  &  317  &  0.78  &  315  &  0.84  &  317 \\
DAYTON\_hourly                 &  5.81  &  370  &  3.91  &  327  &  3.91  &  406  &  2.80  &  327  &  2.72  &  330 \\
DEOK\_hourly                   &  5.38  &  313  &  2.03  &  322  &  2.03  &  348  &  2.84  &  336  &  1.91  &  334 \\
DM                             &  3.39  &  321  &  2.38  &  321  &  3.55  &  317  &  3.05  &  339  &  2.32  &  359 \\
\bottomrule
    \end{tabular}
\end{table}

\begin{table}[ht]
    \centering

    \scriptsize
    \setlength\tabcolsep{3.5pt}

    \caption{Raw experiment results for all evaluated frameworks in the ablation study and used datasets using RMSE. Predictions with consistent failures are marked with --.}
    
    \begin{tabular}{@{} l  c c  c c  c c  c c  c c @{}}
    \toprule
Dataset & \multicolumn{2}{c}{Auto-PyTorch} & \multicolumn{2}{c}{Multi-Fidelity} & \multicolumn{2}{c}{Templates} & \multicolumn{2}{c}{Warm-Starting} & \multicolumn{2}{c}{auto-sktime} \\
        & \rotatebox[origin=c]{90}{RMSE}&\rotatebox[origin=c]{90}{Duration} & \rotatebox[origin=c]{90}{RMSE}&\rotatebox[origin=c]{90}{Duration} & \rotatebox[origin=c]{90}{RMSE}&\rotatebox[origin=c]{90}{Duration} & \rotatebox[origin=c]{90}{RMSE}&\rotatebox[origin=c]{90}{Duration} & \rotatebox[origin=c]{90}{RMSE}&\rotatebox[origin=c]{90}{Duration} \\
\midrule

AEDemand                       &  --    &  310  &  0.48  &  320  &  0.48  &  311  &  0.49  &  309  &  0.04  &  307 \\
AEP\_hourly                    &  2807.39  &  317  &  1758.63  &  341  &  2074.84  &  416  &  1222.00  &  320  &  1222.00  &  323 \\
AirPassengers                  &  24.15  &  321  &  26.08  &  323  &  25.89  &  317  &  22.50  &  315  &  23.84  &  317 \\
DAYTON\_hourly                 &  442.30  &  370  &  317.72  &  327  &  317.72  &  406  &  225.38  &  327  &  219.94  &  330 \\
DEOK\_hourly                   &  607.46  &  313  &  255.92  &  322  &  255.92  &  348  &  332.32  &  336  &  231.93  &  334 \\
DM                             &  0.10  &  321  &  0.07  &  321  &  0.10  &  317  &  0.09  &  339  &  0.07  &  359 \\
\bottomrule
    \end{tabular}
\end{table}

\begin{table}[ht]
    \centering

    \scriptsize
    \setlength\tabcolsep{3.5pt}

    \caption{Raw experiment results for all evaluated frameworks in the ablation study and used datasets using sMAPE. Predictions with consistent failures are marked with --.}
    
    \begin{tabular}{@{} l  c c  c c  c c  c c  c c @{}}
    \toprule
Dataset & \multicolumn{2}{c}{Auto-PyTorch} & \multicolumn{2}{c}{Multi-Fidelity} & \multicolumn{2}{c}{Templates} & \multicolumn{2}{c}{Warm-Starting} & \multicolumn{2}{c}{auto-sktime} \\
        & \rotatebox[origin=c]{90}{sMAPE}&\rotatebox[origin=c]{90}{Duration} & \rotatebox[origin=c]{90}{sMAPE}&\rotatebox[origin=c]{90}{Duration} & \rotatebox[origin=c]{90}{sMAPE}&\rotatebox[origin=c]{90}{Duration} & \rotatebox[origin=c]{90}{sMAPE}&\rotatebox[origin=c]{90}{Duration} & \rotatebox[origin=c]{90}{sMAPE}&\rotatebox[origin=c]{90}{Duration} \\
\midrule

AEDemand                       &  --    &  310  &  0.00  &  320  &  0.00  &  311  &  0.00  &  309  &  0.00  &  307 \\
AEP\_hourly                    &  0.16  &  317  &  0.09  &  341  &  0.11  &  416  &  0.06  &  320  &  0.06  &  323 \\
AirPassengers                  &  0.04  &  321  &  0.05  &  323  &  0.05  &  317  &  0.04  &  315  &  0.04  &  317 \\
DAYTON\_hourly                 &  0.18  &  370  &  0.12  &  327  &  0.12  &  406  &  0.08  &  327  &  0.08  &  330 \\
DEOK\_hourly                   &  0.16  &  313  &  0.06  &  322  &  0.06  &  348  &  0.08  &  336  &  0.05  &  334 \\
DM                             &  0.04  &  321  &  0.03  &  321  &  0.05  &  317  &  0.04  &  339  &  0.03  &  359 \\
\bottomrule
    \end{tabular}
\end{table}

\FloatBarrier

\section{Ranking of Evaluated Methods}

\begin{figure}[h]
    \centering
    \begin{subfigure}[b]{0.45\textwidth}
        \centering
        \includegraphics[width=\textwidth]{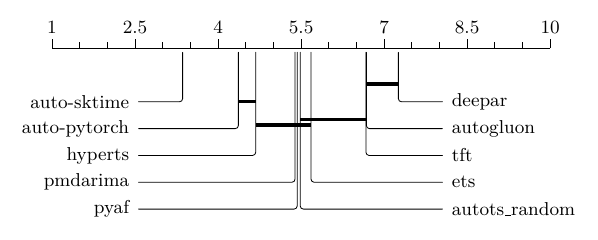}
        \caption{All time series while enforcing timeouts.}
    \end{subfigure}
    \hfill
    \begin{subfigure}[b]{0.45\textwidth}
        \centering
        \includegraphics[width=\textwidth]{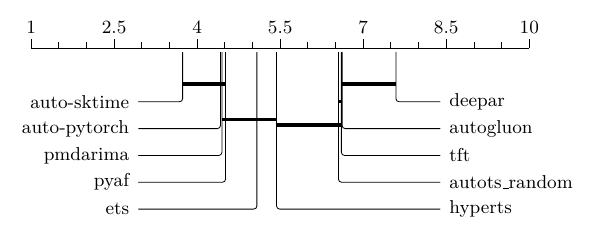}
        \caption{Time series without missing values with timeouts.}
    \end{subfigure}

    \begin{subfigure}[b]{0.45\textwidth}
        \centering
        \includegraphics[width=\textwidth]{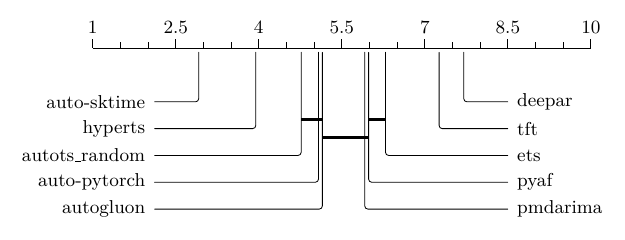}
        \caption{All time series without enforcing timeouts.}
    \end{subfigure}
    \hfill
    \begin{subfigure}[b]{0.45\textwidth}
        \centering
        \includegraphics[width=\textwidth]{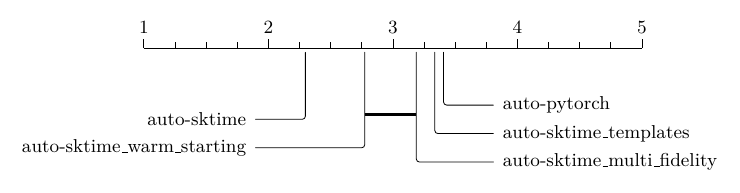}
        \caption{All time series while enforcing timeouts.}
    \end{subfigure}
    \caption{Critical difference diagram of all evaluated framework combinations.}
    \label{fig:critical_difference}
\end{figure}

Figure~\ref{fig:critical_difference} contains critical difference plots for all evaluated scenarios in Section~4.2 and 4.3. The average ranking of each evaluated method is displayed, and whether observed values differ significantly from each other according to a \(t\)-test with significance level \(p = 0.05\) and Bonferroni correction for multiple hypothesis tests.

\section{Used Computing Hardware}

All experiments were conducted on standard hardware, namely \emph{e2-standard-4} virtual machines on Google Cloud Platform equipped with Intel Xeon E5 processors with four cores and 16 GB memory. Even though multiple tested frameworks construct neural networks, no graphics card was used. Some frameworks make extensive usage of parallelization of workload over all available cores, while other frameworks do not do this. To ensure fair comparisons, all frameworks were limited to exactly one CPU core to ensure identical computational budgets.